%% file: acl2023.tex
\documentclass[11pt]{article}

\usepackage[utf8]{inputenc}
\usepackage[most]{tcolorbox}
\usepackage{graphicx} 
\usepackage{authblk}
\usepackage{ACL2023}
\usepackage{amsmath} 
\usepackage{booktabs} 
\usepackage{times}
\usepackage{latexsym}
\usepackage{graphicx}
\usepackage{multirow} 
\usepackage{enumitem}

\usepackage[T1]{fontenc}

\usepackage{float}
\usepackage{afterpage}
\usepackage{caption}
\usepackage[utf8]{inputenc}

\usepackage{microtype}

\usepackage{inconsolata}
\title{SG-FSM: A Self-Guiding Zero-Shot Prompting Paradigm  for Multi-Hop Question Answering Based on Finite State Machine}

\author[1,2,3]{Xiaochen Wang }  \author[1]{Liang Chen} \author[4]{Reza Haf} \author[1]{Zhe Yang} \author[5]{Yiru Wang} \author[1,2]{\\ Xiangdi Meng}  \author[3]{Kunhao Pan} \author[3]{Jiaxing Zhang} \author[1,$\dagger$]{ Zhifang Sui } \author[3,$\dagger$]{Junqing He}

\renewcommand*{\Affilfont}{\small\it} 

\affil[1]{National Key Laboratory for Multimedia Information Processing, Peking University}

\affil[2]{School of Software \& Microelectronics, Peking University}
\affil[3]{International Digital Economy Academy}

\makeatletter
\renewcommand\AB@affilsepx{, \protect\Affilfont}  
\makeatother

\affil[4]{Data Science and Artificial Intelligence, Monash University}
\affil[5]{ModelTC}

\begin{document}
\maketitle

\begin{abstract}

Large Language Models with chain-of-thought prompting, such as OpenAI-o1, have shown impressive capabilities in natural language inference tasks. However, Multi-hop Question Answering (MHQA) remains challenging for many existing models due to issues like hallucination, error propagation, and limited context length. To address these challenges and enhance LLMs' performance on MHQA, we propose the Self-Guiding prompting Finite State Machine (SG-FSM), designed to strengthen multi-hop reasoning abilities. Unlike traditional chain-of-thought methods, SG-FSM tackles MHQA by iteratively breaking down complex questions into sub-questions, correcting itself to improve accuracy. It processes one sub-question at a time, dynamically deciding the next step based on the current context and results, functioning much like an automaton. Experiments across various benchmarks demonstrate the effectiveness of our approach, outperforming strong baselines on challenging datasets such as Musique. SG-FSM reduces hallucination, enabling recovery of the correct final answer despite intermediate errors. It also improves adherence to specified output formats, simplifying evaluation significantly.

\end{abstract}

\section{Introduction}

Multi-hop Question Answering (MHQA) is a challenging QA task that asks models to answer a complex and indirect question given multiple passages. Agents need to reason twice/more on documents to get the final answer. It has intrigued researchers for its complexity and practical implications \cite{2wiki,hotpot,musi}. 

Researchers employ three primary strategies to address MHQA using Large Language Models (LLMs) due to their powerful and promising  ability. 
One effective method is In-Context Learning (ICL) \cite{self-prompted,least_to_most,react}, where models are instructed to solve problems based on detailed demonstrations. However, few-shot methods are considered ineffective and inefficient as they require a minimum of 4-shot of manual designed demonstrations. Long context in these demonstrations may exceed context boundaries and distract attention \cite{lostinmiddle}. Another approach involves fine-tuning LLMs with domain-specific data, which requires substantial high-quality data and computational resources. It is effective but inefficent and non-generalizable. 
The third method reduces the training cost by training a new module for only part of the procedure without training LLMs. For example, \cite{train_kuaishou} trained a beam-retrieval model. After retrieving results, they utilize a few-shot LLM as a reader to answer the question. However, they only improved the system's retrieval capabilities, while the LLM still exhibits hallucinations and propagates errors.

In this paper, we summarize four common inference errors in the previous approaches and demonstrate the effectiveness of our method in tackling these issues. We found that LLMs struggle particularly in intermediate reasoning stages, where errors in initial steps can propagate, leading to incorrect conclusions. We define this error as lost-in-the-middle reasoning path. 
Detailed analysis is presented in the \S~\ref{Sec:Disscusion}.

\begin{figure*}[ht]
  \centering
  \includegraphics[width=\textwidth,  height=0.31\textheight]{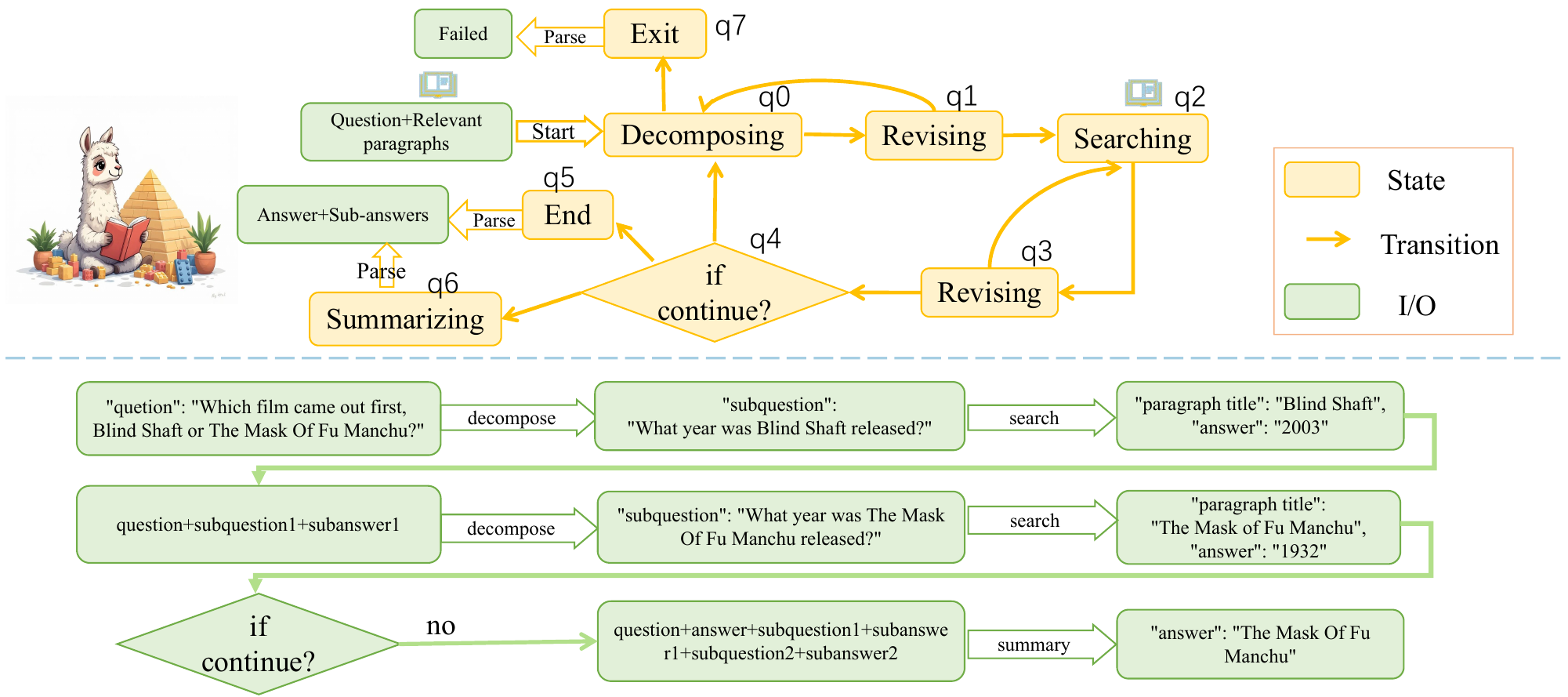}
  \caption{ The flow chart of proposed SG-FSM. The upper part illustrates the flow of SG-FSM with state transitions. $q_i$ is state defined in five-tuples. The lower part shows the input/output flows of an example.} 
  \label{fig_automat}
\end{figure*}

Humans usually decompose sophisticated problems to solve them, supported by cognitive discoveries \cite{correa2023humans,cheng2015break}. Many decomposition methods to assist LLMs have shown their effectiveness in other tasks \cite{decomposing-complex,he-etal-2024-never}. Inspired by the insights, this paper adapts the decomposing progress to MHQA to improve the performance. Herein we decompose an MHQA task in predefined order: first, identify the initial sub-question, then search for its answer in the text, and continue solving each subsequent sub-question in sequence until the final answer is reached. The process is similar to a Finite State Machine (FSM), which can constrain the intermediate reasoning process and shorten the lengthy reasoning path.

According to the analysis above, we propose a zero-shot method named \textbf{S}elf-\textbf{G}uiding \textbf{F}inite \textbf{S}tate \textbf{M}achine prompting (SG-FSM), simplifying the MHQA task into four sub-tasks: decomposing questions, searching for answers in candidate paragraphs, revising the format, and judging whether to continue. These four tasks resemble states in SG-FSM, and SG-FSM loops through these sub-task states sequentially until the final answer is found. Lastly, SG-FSM summarizes key information ahead and self-corrects. Figure \ref{fig_automat} depicts the process of the SG-FSM. 
We declare the advantages of SG-FSM to REAC as follows. In REACT, the 'control' over different stages is straightforward, with transitions following a chain-like structure across the states (ie the next step CoT, call the API, get the answer, go to the next step CoT). Whereas in SG-FSM, the control transitions across a fine-state-automata over the states, adaptively selecting the next state based on the results of the previous steps.

\begin{figure*}[ht!]
  \centering
 \includegraphics[width=0.99\linewidth, keepaspectratio]{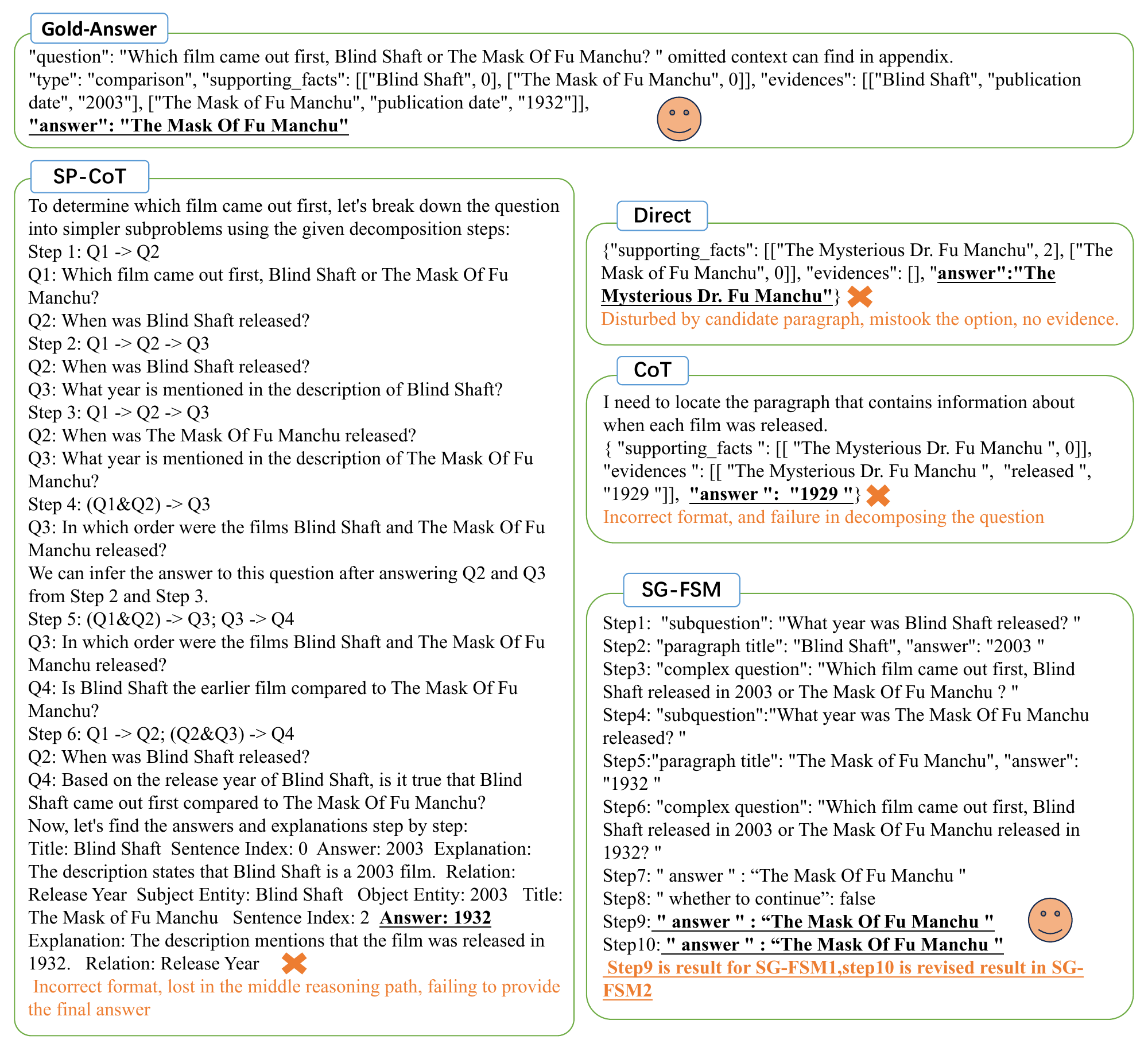}
  \caption{Outputs for different methods. Each error is marked, and SG-FSM can solve these errors.}
  \label{fig_error}
\end{figure*}

Extensive experiments on MHQA benchmarks \cite{hotpot,musi,2wiki} with GPT-3.5-turbo-1106 and Qwen-72B demonstrate that our approach outperforms baselines, nearly doubling the F1 score on Musique \cite{musi}. Baselines often generate outputs in unexpected format errors which are hard to process while SG-FSM greatly reduces the format error and shows robustness. More importantly, SG-FSM also improves the correctness of intermediate reasoning and supporting evidence significantly while those of baselines are faulty even if they give a correct answer, as demonstrated in Figure \ref{fig_error}.

Our contributions are as follows:

\textbullet~ We summarize common error cases in MHQA and analyze the reasons. Besides, we discover an unexpected phenomenon in the long free reasoning path using COT, which we called "lost-in-the-middle reasoning path". 

\textbullet~We introduce SG-FSM, a self-guiding zero-shot prompting paradigm to decompose complex questions iteratively, enhancing the capability of LLMs to solve complex problems through a controlled reasoning path.

\textbullet~Extensive experiments on MHQA benchmarks in different settings validate SG-FSM's effectiveness, especially on challenging datasets.

\section{Preliminary} 
\label{Sec:Disscusion}
Figure~\ref{fig_error} presents error examples from the baseline methods and highlights the advantages of the proposed SG-FSM approach. Specifically, the direct method struggled with the long candidate text sequence, leading to errors in selecting the correct option and failing to provide supporting evidence. The CoT method did not produce the required JSON format and was unable to effectively decompose the problem, addressing only one sub-question. SP-COT repeatedly decomposed the problem but lost track of the original question, ultimately failing to deliver the final answer after comparison. In contrast, SG-FSM correctly inferred each step, arriving at the accurate final answer.
After analyzing extensive cases, we concluded four primary reasons for bad cases in MHQA as follows:

\textbf{a) Error Propagation:} CoT is prone to introducing errors during intermediate reasoning steps, such as mistakes in decomposition and searching. These errors accumulate and propagate through the reasoning process, a phenomenon known as Rationale Drift in \cite{cotpoorexplain}.

\textbf{b) Lost-in-the-middle reasoning path: }Extended context in the reasoning process can lead the LLM to lose focus on the original question, making it difficult to provide the correct answer. It is called Answer Drift in \cite{cotpoorexplain}, when the model loses track of the question during inference. 

\textbf{c) Format Mismatch:} Correct answers may not be recognized during evaluation due to format errors. It fails to follow instructions probably due to long contexts or hallucinations. Examples are presented in Section ~\ref{format_error}. 

\textbf{d) Hallucination Response}: Provided a correct answer without locating the relevant paragraph. 

Issues of unfaithfulness and self-contradiction in the LLMs reasoning process are emerging areas. \citet{measuring_faithfulness,score,Self-contradictory} point out that in existing reasoning research, much work overly focuses on the predictive accuracy of models, neglecting the quality and consistency of the reasoning process itself.

\section{Methodology}
\subsection{Task Definition}

The issue of multi-hop QA is characterized by a question $q$ and a set of pertinent (gold) supporting context documents $d_1, \ldots, d_S$ that hold the answer $a$. These context documents create a logical sequence essential for reaching the answer, drawn from a vast collection of documents $D$ where the size of $D$ greatly exceeds $S$. Given a multi-hop question and multiple related paragraphs, the model is required to provide the final answer, as well as the paragraph location to find the sub-answers for sub-questions. 


\subsection{Framework}

We present our proposed Self-Guided Finite State Machine (SG-FSM) in two distinct stages as illustrated in Figure ~\ref{fig_automat}. Initially, we instruct LLMs to address sub-questions iteratively during the first phase, SG-FSM1 for short. Subsequently, in stage 2, LLMs are asked to summarize key information from each sub-question and instruct it to revise based on results of SG-FSM1. This is the difference between SG-FSM1 and SG-FSM2.  

The SG-FSM is formally described as a five-tuple $(Q, \Sigma, \delta, q_0, F)$, where:

\begin{itemize}[itemsep=-4pt]
    \item $Q = \{q_0, q_1, q_2, q_3, q_4, q_5\}$ is the set of states, where:
    \begin{itemize}
        \item $q_0$: Decomposing the question
        \item $q_1$: Revising the output of decomposing
        \item $q_2$: Searching in the given paragraph
        \item $q_3$: Revising the output of searching
        \item $q_4$: Judging if question can be decomposed further
        \item $q_5$: The end of SG-FSM1, final answer is found 
        \item $q_6$: Summarizing with key reasoning information
        \item $q_7$: Early withdrawal
               
    \end{itemize}
    \item $\Sigma = \{\text{question}\}$ is the complex question, representing the input to the system
    \item $\delta: Q \times \Sigma \rightarrow Q$ is the transition function, defined as follows:
    
      $\delta(q_0, \text{question}) = \begin{cases} 
            q_1, \text{if correct} \\            
            q_7, \text{elif iterations}>6 \\
            q_0, \text{else}
        \end{cases}$
      
         $\delta(q_1, \text{output in }  q_0) = 
        \begin{cases} 
            q_2, \text{if correct} \\
            q_0, \text{else}
        \end{cases}$
        \\
        \\
         $\delta(q_2, \text{paragraph}) = q_3$
         \\
         \\
         $\delta(q_3, \text{output in } q_2) = 
        \begin{cases} 
            q_4 & \text{if correct} \\
            q_2 & \text{else}
        \end{cases}$
        \\
         $\delta(q_4, \text{history}) = 
        \begin{cases} 
            q_0 & \text{if continue} \\
            q_5 & \text{else}
        \end{cases}$
        \\
        \\
         $\delta(q_5, \text{paragraph}) = q_6$

    Where "if correct" means output can be parsed correctly, and "if continue" indicates whether the question can be further decomposed. 
    \item $q_0$ is the initial state
    \item $F = \{q_5, q_6, q_7\}$ is the set of accept states, indicating the terminal state where the final answer is found or early exit.
\end{itemize}

The specific finite state automaton diagram is located in the upper part of Figure ~\ref{fig_automat}. It shows clear process of the FSM and how the states are transitioned. 
As the problem-solving steps for MHQA tasks adhere to a consistent pattern, they can be classified into four clear phases, with the model concentrating on addressing one task sequentially. These components are outlined as Decomposer, Searcher, Revisor, Terminator and Summarizer. All of them are performed by the same LLM, each utilizing a different prompt to ensure that only one task is processed in each round. Combining the four tasks may introduce more complexity than performing them individually, as a single error in reasoning could lead to the failure of the entire task. 
The specific prompt can be found in the Appendix ~\ref{prompt}. The components will be described in detail in the following subsections with the inputs and outputs of each being presented.

An example is depicted in Figure ~\ref{fig_automat}. Let's describe its specific steps in detail. When parsing the output results at each step, if there are formatting errors or other issues, it is necessary to immediately use revision to correct output. If two attempts fail, SG-FSM1 will exit the loop early. Therefore, we omit revision steps for brevity. 

First, the decomposer breaks down the original question and gives one sub-question \textit{"What year was Blind Shaft released?"}. In the next turn, the searcher finds its answer in the multiple candidate paragraphs and outputs the referred paragraph title and answer. Since the first sub-question is addressed, it comes to the terminator to judge if the question can be decomposed further. Now the complex question actually becomes \textit{"Which film came first, Blind Shaft 
 released in 2003 or The Mask Of Fu Manchu?"}. Obviously, the SG-FSM1 should continue.
 Then, decomposer gives another sub-question \textit{"What year was The Mask Of Fu Manchu released?"}. Then the searcher finds the answer "1932" in the reference paragraph. Next, the terminator answers the question \textit{"Which film came first, Blind Shaft 
 released in 2003 or The Mask Of Fu Manchu released in 1932?"} and quits this loop. The answer is \textit{"The Mask Of Fu Manchu"}.
 
 In the SG-FSM2 stage, we give LLMs key reasoning information ahead and instruct LLMs to revise these and answer the question again.

\newenvironment{prompt}
    {\begin{tcolorbox}[
        enhanced,
        title=prompt,
        fonttitle=\bfseries,
        colback=white,
        colframe=blue!60!black,
        colbacktitle=blue!20!white,
        attach boxed title to top center={yshift=-3mm,yshifttext=-1mm},
        boxrule=0.3mm, 
        titlerule=0.3mm 
        toprule=0.3mm, 
        leftrule=0.3mm, 
        rightrule=0.3mm, 
        fontupper=\small 
    ]}
    {\end{tcolorbox}}

\newcommand{\infoitem}[2]{\textbf{#1:} #2\par}

\newenvironment{promptd}
    {\begin{tcolorbox}[
        enhanced,
        title=I/O for Decomposer,
        fonttitle=\bfseries\small ,
        colback=white,
        colframe=blue!60!black,
        colbacktitle=blue!20!white,
        coltitle=black, 
        attach boxed title to top center={yshift=-3mm,yshifttext=-1mm},
        boxrule=0.3mm, 
        titlerule=0.3mm 
        toprule=0.3mm, 
        leftrule=0.3mm, 
        rightrule=0.3mm, 
    ]}
    {\end{tcolorbox}}

\newcommand{\infoitemd}[2]{\textbf{#1:} #2\par}

\subsection{Decomposer}
We need to ensure that the LLMs solve the problem step by step, so we decompose the complex problem to make it theoretically easier to answer. The input and output are as follows:
\begin{promptd}

 \infoitemd{Input}{ 
 
 Please determine whether the question is simple sentence or compound sentence. If it is a simple sentence, return \textbraceleft "simple":true,"subquestion":null \textbraceright.Otherwise decompose the question and generate the first answerable simple sentence. 

 Reply in the form of \textbraceleft "simple":false, "subquestion":xxx \textbraceright. + Requirements.

 Question: "Which film came first, Blind Shaft or The Mask Of Fu Manchu?"} 

  \infoitemd{Output}{  \textbraceleft"simple": false, "subquestion": What year was The Mask Of Fu Manchu released?\textbraceright}
  
\end{promptd}

To instruct LLM to output compliant JSON format for convenient parsing, the requirements should be strictly defined:
\textit{\textbraceleft examples of output format\textbraceright. Do not reply any other words and provide answers in JSON format!}
The output requirements in subsequent prompts are similar; for brevity, they will be omitted.

\subsection{Searcher}
Given sub-question ahead and candidate paragraph in origin task, searcher will find the answer  (and supportting evidence in setting 2) directly.
\newenvironment{prompts}
    {\begin{tcolorbox}[
        enhanced,
        title=I/O for Searcher,
        fonttitle=\bfseries,
        colback=white,
        colframe=blue!60!black,
        colbacktitle=blue!20!white,
        coltitle=black, 
        attach boxed title to top center={yshift=-3mm,yshifttext=-1mm},
        boxrule=0.3mm, 
        titlerule=0.3mm 
        toprule=0.3mm, 
        leftrule=0.3mm, 
        rightrule=0.3mm, 
    ]}
    {\end{tcolorbox}}

\newcommand{\infoitems}[2]{\textbf{#1:} #2\par}

\begin{prompts}
 \infoitems{Input}{

 Given the paragraph below, please find out the paragraph that contains the answer of "\textbraceleft\textbraceright" Please take a moment to thoroughly understand the content before proceeding to the questions, then carefully read the relevant paragraphs based on the question and provide the most likely answer. 
 \\
 Question: "What year was The Mask Of Fu Manchu released?
 \\
 Context: paragraph...
 } 
  \infoitems{Output}{
  The answer is \textbraceleft \""subanswer: 1932,   }
 
\end{prompts}

\subsection{Revisor}
After each step, the LLMs output content should be immediately parsed for analysis, and any errors should be corrected immediately. Only outputs with format errors enter this revisor step to self-correction. If there is still an error after two retries, we terminate the loop early and mark the answer as blank.
\newenvironment{promptr}
    {\begin{tcolorbox}[
        enhanced,
        title=I/O for Revisor,
        fonttitle=\bfseries,
        colback=white,
        colframe=blue!60!black,
        colbacktitle=blue!20!white,
        coltitle=black, 
        attach boxed title to top center={yshift=-3mm,yshifttext=-1mm},
        boxrule=0.3mm, 
        titlerule=0.3mm 
        toprule=0.3mm, 
        leftrule=0.3mm, 
        rightrule=0.3mm, 
    ]}
    {\end{tcolorbox}}

\newcommand{\infoitemr}[2]{\textbf{#1:} #2\par}

\begin{promptr}
\infoitemr{Input}{ 

Please rewrite the illegal json text below into an legal json string.
Text: The answer is \textbraceleft \""subanswers": \" 1932,\textbraceright  }
\infoitemr{Output}{
\textbraceleft "subanswer": 1932\textbraceright }
\end{promptr}
\subsection{Terminator}
Currently, most MHQA questions require 2-4 hops of reasoning. After addressing a sub-question, we need to determine whether the question has been fully decomposed. If the final answer has been discovered, exit this loop and proceed to the final SG-FSM2 summary stage.
\newenvironment{promptt}
    {\begin{tcolorbox}[
        enhanced,
        title=I/O for Terminator,
        fonttitle=\bfseries,
        colback=white,
        colframe=blue!60!black,
        colbacktitle=blue!20!white,
        coltitle=black, 
        attach boxed title to top center={yshift=-3mm,yshifttext=-1mm},
        boxrule=0.3mm, 
        titlerule=0.3mm 
        toprule=0.3mm, 
        leftrule=0.3mm, 
        rightrule=0.3mm, 
    ]}
    {\end{tcolorbox}}

\newcommand{\infoitemt}[2]{\textbf{#1:} #2\par}

\begin{promptt}
\infoitemt{Input}{

Can the original question be further decomposed into other different sub-question? Please output in the form of \textbraceleft"continue":true or false\textbraceright.
\\
original question: "Which film came first, Blind Shaft or The Mask Of Fu Manchu?
 sub-question: "What year was The Mask Of Fu Manchu released?}
\infoitemt{Output}{\textbraceleft "continue": true\textbraceright }
\end{promptt}

\subsubsection{Summarizer}
The previous modules together form SG-FSM1. Although the SG-FSM1 phase has already generated the answer, it may still contain logical errors, so we add SG-FSM2, listing all the key information and letting the LLMs check it again. 
\newenvironment{promptsu}
    {\begin{tcolorbox}[
        enhanced,
        title=I/O for Summarizer,
        fonttitle=\bfseries,
        colback=white,
        colframe=blue!60!black,
        colbacktitle=blue!20!white,
        coltitle=black, 
        attach boxed title to top center={yshift=-3mm,yshifttext=-1mm},
        boxrule=0.3mm, 
        titlerule=0.3mm 
        toprule=0.3mm, 
        leftrule=0.3mm, 
        rightrule=0.3mm, 
    ]}
    {\end{tcolorbox}}

\newcommand{\infoitemsu}[2]{\textbf{#1:} #2\par}

\begin{promptsu}
\infoitemsu{Input}{ 

Original question: Which film came first, Blind Shaft or The Mask Of Fu Manchu?
 \\
 Sub-question 1: What year was The Mask Of Fu Manchu released?
\\    Paragraph: The Mask of Fu Manchu...
\\    Evidence: (The Mask of Fu Manchu, released in, 1932)
 \\   Sub-answer: 1932
 \\   
   Sub-question 2: What year was Blind Shaft released?
 \\   Paragraph: Blind Shaft...
\\    Evidence: (Blind Shaft, released in, 2003)
\\    Sub-answer: 2003
    Answer: The Mask of Fu Manchu
\\
Please check based on the above information whether each sub-question's answer is correct, and whether the given answer is correct to the original question. Output the final correct answer in the form of \textbraceleft"Answer": xxx, "Reason": xxx\textbraceright.."}    
\infoitemsu{Output}{
\textbraceleft "Answer": The Mask of Fu Manchu,"Reason": ...\textbraceright }
\end{promptsu}

\section{Experiments}
\subsection{Benchmark and Evaluation}
\label{benchmark}
We evaluate models on three high-quality MHQA datasets: HotpotQA \cite{hotpot},  2WikiMultiHopQA (2WikiQA) \cite{2wiki} and Musique \cite{musi}. Learning from the shortcut phenomenon \cite{Min_no_need_mhreason} of single hop questions in HotpotQA, Musique \cite{musi} strictly controls the composition of the question, ensuring it undergoes multiple inferences to find the answer. Both HotpotQA and 2WikiQA have ten candidate paragraphs for each question and originally have supporting facts. In comparison, Musique has twenty candidates with longer text and no supporting facts. Question composition types and number of hops are listed in Table \ref{statistics}. More question types and hops are contained in Musique. Therefore, Musique is the most difficult MHQA dataset. 
\begin{table}[h]
    \centering
    \small
    \begin{tabular}{c|ccc}
    \toprule
        Datasets  & Hotpot QA & 2WikiQA & Musique \\ \midrule
        \#Hops  & 1-2 & 2 & 2-4 \\
        \#Types & 2 &4 & 6\\
        \#Paragraphs & 10 &10 & 20\\
    \bottomrule
    \end{tabular}
    \caption{Statistics of datasets in experiments. Type is short for question composition types. Paragraphs represent the number of candidate paragraphs for each sample.}
    \label{statistics}
\end{table}

Following \cite{self-prompted}, we adopt the exact match (EM) and F1 scores as evaluation metrics and conduct experiments on subsets of the datasets by randomly selecting 1000 samples from the test sets. Despite having similar basic instructions and a clearly defined output format for all methods, the model's consistency in following instructions may vary across different methods. This variation can result in difficulty in answer extraction during evaluation. To address this issue, we introduce a new metric, format, measuring the accuracy of the output format.
\input{floats/setting1}

\input{floats/bigdata}
\subsection{Baselines}
We conduct experiments by considering the following baselines with both open-sourced and API endpoints:

\textbullet~The \textbf{Direct} strategy inference the answer directly, which is the basic form, using only task descriptions and output requirements as the prompt.

\textbullet~The \textbf{CoT} \cite{cot} prompts LLMs to create intermediate step-by-step rationales, aiding in the reasoning process for obtaining answers.

\textbullet~The \textbf{SP-CoT} \cite{self-prompted} organizes reasoning chains into six categories, inspired by the construction of the Musique \cite{musi} dataset. It designs multiple demonstrations and then selects the suitable ones for in-context learning.

All the prompts for the baselines above are included in the Appendix ~\ref{prompt}.

\subsection{Settings}
Our study explores two settings: (1) only asks for answers given the context and question without the need for supporting facts, and (2) building a complete reasoning chain that includes the answer, supporting evidence, and facts to assess the coherence of the reasoning process. Due to the lack of golden evidence for Setting 2 in Musique, our evaluation did not report its results. Existing methods mostly adopt Setting1 and do not report Setting 2. By adding Setting 2, we can observe how enhancing the inference process affects the output of the LLMs, highlighting instances where the model provides the correct answer through an incorrect intermediate process.

\subsection{Models}
Since MHQA requires models with the ability to process lengthy text for multiple rounds of reasoning, we selected GPT-3.5-turbo-32k \footnote{https://platform.openai.com/docs/models/gpt-3-5-turbo} and Qwen72B-chat \cite{qwen} for our study. Additionally, we employed Vllm \cite{vllm} to accelerate the inference process.

\subsection{Results}
The results for Setting 1 (sole answer) are presented in Table \ref{tab:setting1}, while the outcomes for Setting 2 (answers paired with supporting facts) are shown in Table \ref{tab:setting2}. 
Our method generally outperforms the baseline, showcasing the effectiveness of SG-FSM. By dissecting the question step by step, we enhance the accuracy of each step towards the sub-questions. This focused approach during searching ensures that attention is not diverted by extraneous texts, leading to more precise results.

Notably, SG-FSM excels on the Musique dataset, showing a significant improvement of 6-10 percentage points. This dataset, as discussed in Section ~\ref{benchmark}, poses more hops and longer reasoning paths, making it more challenging than others. Our method provides clear guidance to LLMs at each step, promptly verifying the actions taken, thus easing the task complexity. Additionally, its performance surpasses that of HotpotQA and 2WikiQA, as it avoids single-hop shortcuts \cite{Min_no_need_mhreason}, further validating the effectiveness of our method.


Besides, the direct method struggles with supporting facts but achieves substantially higher scores on answers than CoT. This phenomenon indicates that although LLMs may misinterpret intermediate reasoning steps, they still yield correct answers, hinting at underlying data leakage and hallucination. While some errors may stem from misinterpreting instructions, significant concerns regarding the authenticity and logical coherence of the models' reasoning chains. 

Because of severe hallucination issues and data leakage problems in the LLMs, we call for a new benchmark that uses GPT to evaluate the logical consistency throughout the entire reasoning process. Under this benchmark, the effectiveness of our method can be better demonstrated.


\subsection{Ablation Study}
In our previous case analysis, we observed instances of hallucination and deceptive reasoning in LLMs when responding to questions: they provided correct answers but used erroneous reasoning. To investigate this phenomenon further, we compared the preliminary results obtained before the final summary revision to those after the revision, specifically contrasting SG-FSM1 with SG-FSM2.

Generally, the results of SG-FSM2 are more accurate compared to SG-FSM1, although its scores fluctuate. In one scenario, if answer is correct, relevant documents are incorrect, then after correction, the answer becomes unavailable. This corrects the illusion, and the score will decrease. In the other scenario, if relevant documents are correct but answer is wrong, and only then the individual corrects the answer to be correct, the score will increase. In summary, SG-FSM2 can only provide the correct answer when the appropriate relevant documents are supplied. The case where SG-FSM2 is worse than SG-FSM1, correct answer and wrong intermediate reasoning evidence, indicates that LLMs suffers from severe hallucination issues.

\section{Related Work}

\textbf{Multi-hop QA datasets} 
HotpotQA \cite{hotpot} is the most classic dataset. Subsequently, 2WikiMultiHopQA \cite{2wiki} was proposed. Learning from the shortcut phenomenon \cite{Min_no_need_mhreason} of single hop questions in HotpotQA, Musique strictly controls the composition of the question, ensuring that it must undergo multiple inferences to find the answer.

\subsection{Multi-hop Question Answering} 

Existing approaches to solving the multi-hop QA task can be mainly categorized into  question decomposition \cite{decompose_trained,decomposing-complex,decompse_unsupervise}, graph-based method \cite{gnn-hgn,gnnidentifying,gnnmhqa}, iterative method \cite{iterly_q} and LLMs \cite{self-prompted} prompts. These models grapple with computational complexity and extensibility, and they lack an interpretable reasoning chain, which deviates from human cognitive processes.

\subsection{Lage Language model for reasoning.} 

CoT\cite{cot} reveals the ability of large language models to formulate their reasoning procedure for problem-solving. Several follow-up works have since been performed, including the least-to-most prompting technique \cite{least_to_most} for solving complicated tasks, zero-shot CoT \cite{zero-cot}, graph-of-thought (GoT) \cite{got}, and reasoning with self-consistency \cite{self-consistency}. ReAct \cite{react} interleaves the generation of reasoning traces with task-specific actions, promoting greater synergy. Recently, OpenAI-o1 \footnote{https://openai.com/o1/} series models perform remarkably well, as they are trained with reinforcement learning to execute complex reasoning tasks. The key feature of OpenAI-o1 is its methodical approach, generating a long internal chain of thought before responding to user queries. Our work proved the necessity of extending and splitting the inference chain before it.

\subsection{Task decomposition.} 
\citet{decompose_trained} decomposes a multi-hop question into a number of independent single-hop sub-questions, which are answered by an off-the-shelf question-answering (QA) model. These answers are then aggregated to form the final answer. Both question decomposition and answer aggregation require training models. After the emergence of Large Language Models (LLMs), traditional training methods \cite{train_kuaishou} are rarely used due to their expensive nature. Most current research focuses on the few-shot approach. \cite{least_to_most} chains the processes of problem decomposition and sub-problem solving. The original problem and its sub-problems are inherently interrelated, and forcibly breaking them down into unrelated problems would unnecessarily increase the difficulty.

\section{Conclusion}
We investigated and classified error reasons in traditional methods where LLMs underperform the MHQA task in the paper. Besides, we discover an unexpected phenomenon in the long free reasoning path using CoT, called "lost-in-the-middle reasoning path".
To address these issues, we propose SG-FSM, a self-guiding zero-shot prompting approach to break down intricate questions step by step iteratively. This method improves the ability of LLMs to tackle difficult problems by guiding them through a controlled and extended reasoning process. Extensive experiments on multiple benchmarks show the superiority of SG-FSM over strong baselines and its effectiveness.
SG-FSM delivers more robust and explainable reasoning output including answers and supporting facts by guiding the reasoning process and performing timely revisions.



\section*{Limitations}
This multi-turn dialogue process, inherent to our framework, mandates repeated handling of improperly formatted outputs, due to the output before will be the next input, which can be challenging for models with smaller parameter sizes and weaker follow-instruction capabilities. Therefore, models with limited capacity to follow instructions might not benefit from our method as any error in the intermediate steps could lead to an abrupt termination of the process.

The primary factor is the LLM's capability, with prompts playing a supporting role. LLMs exhibit varying abilities largely dependent on how they are trained. We believe that the performance of a model is mostly influenced by whether the testing and training data distributions are consistent. Thus, if the model uses our method to incorporate step-by-step inference of control during training, the effect will be better.

\bibliography{custom}
\bibliographystyle{acl_natbib}

\appendix
\section*{Appendix}
\label{sec:appendix}

\section{Prompt}
\label{prompt}
\subsection{SG-FSM1}
Decomposer:
Please determine whether the question is simple sentence or compound sentence. If it is a simple sentence, return {"simple":true,"subquestion":null}.Otherwise, simple: false, decompose the question and generate the first answerable simple sentence. reply in the form of {"simple":false,"subquestion":xxx}. Do not reply any other words and provide answers in JSON format!

Searcher:
Given the paragraph below, please find out the paragraph that contains the answer of "{}" Please take a moment to thoroughly understand the content before proceeding to the questions, then carefully read the relevant paragraphs based on the question and provide the most likely answer. Return the title of the paragraph and the answer no more than 5 words in the form of {"question":xxx, "paragraph title":xxx, "answer":xxx}. Do not reply any other words and provide answers in JSON format!

Judge-if-continue:
Please compare the complex question and subquestion, answer whether they are semanically identical in the form of {"identical":true or false}. Do not reply any other words and provide answers in JSON format!

\subsection{SG-FSM2}
FSM2-post-summary-again:
Documents:
paragraphs:{paragraphs found in FSM1}
subquestion and answers:{subquestion and answers given in FSM1}
Question:{origin question}
Answer the question reasoning step-by-step based on the Doucments. If it is a general question, please respond with 'Yes' or 'No'. Finally, you must return the title of the context, the sentence index (start from 0) of the paragraph and the concise answer no more than 10 words and explaination in the form of {"supporting-facts": [[title, sentence id], ...], "evidences": [[subject entity, relation, object entity],...], "answer":"xxx","explain":"xxxx"}. Do not reply any other words. 

\subsection{Baseline}
SP-CoT\cite{self-prompted}:
This is a two-hop to four-hop reasoning question-answering task that requires decomposing the questions into simple, answerable single-hop questions. The decomposition process involves four types of questions: comparison, inference, compositional, and bridge-comparison. There are six specific decomposition steps in total, denoted by Q representing the decomposed subproblems. The steps are as follows:
First, Q1 -> Q2
Second, Q1 -> Q2 -> Q3
Third, Q1 -> Q2 -> Q3
Fourth, (Q1\&Q2) -> Q3
Fifth, (Q1\&Q2) -> Q3; Q3 -> Q4
Sixth, Q1 -> Q2; (Q2\&Q3) -> Q4
The process involves first determining the type of question and then identifying the decomposition process type. It's important to note that the decomposition of questions cannot be provided all at once; it must be done step by step. Each subproblem needs to be decomposed and answered before moving on to the next one, as there is interdependence between the subproblems .Finally, you must return the title of the context, the sentence index (start from 0) of the paragraph and the concise answer and explaination in the form of 
{"explain":"xxxx","supporting-facts": [[title, sentence id], ...], "evidences": [[subject entity, relation, object entity],...],"answer":"no sentence and no more than 10 words "}. 
Do not reply any other words.

CoT-setting1-w/o-evidence: 
Answer the question according to the context,Let's think step by step, and explain your reasoning process. You must return in the form of {"explain":"xxxx","answer":answer}. Do not reply any other words.

direct-setting1-w/o-evidence:
Answer the question according to the context. You must return in the form of {"explain":"xxxx","answer":answer}. Do not reply any other words.

direct-setting2-w-evidence: 
Answer the question according to the context. Find the paragraph that contains the answer of question, and summarize a triple that contains [subject entity, relation, object entity]. Finally, you must return the title of the context, the sentence index (start from 0) of the paragraph and the concise answer no more than 10 words in the form of {"supporting-facts": [[title, sentence id], ...], "evidences": [[subject entity, relation, object entity],...], "answer":answer}. Do not reply any other words.

prompt-step:  
Answer the question according to the context,Let's think step by step, and explain your reasoning process. Find the paragraph that contains the answer of question, and summarize a triple that contains [subject entity, relation, object entity]. Finally, you must return the title of the context, the sentence index (start from 0) of the paragraph and the concise answer no more than 10 words in the form of {"supporting-facts": [[title, sentence id], ...], "evidences": [[subject entity, relation, object entity],...], "answer":answer}. Do not reply any other words.

React-setting2-w-evidence: 
Solve a question answering task with interleaving Thought, Action, Observation steps. Thought can reason about the current situation, and Action can be three types: 
(1) Search[entity], which searches the exact entity on given context and returns the first paragraph if it exists. If not, it will return some similar entities to search.
(2) Lookup[keyword], which returns the next sentence containing keyword in the current passage.
(3) Finish[results], which returns the answer and finishes the task.
You should plan and reason in the Thought, then perform your Action, lastly, observe the result of action. Loop this process until the problem was finished. 
At last, you must additional output the title of the paragraphs, the sentence index (start from 0) of the paragraph and the concise answer no more than 10 words and explaination in the form of 
Thought: reasoning
Action: Search[entity] or Lookup[keyword] or Finish[results]
Observation: observe the results of action
end with Finish[{"supporting-facts": [[title, sentence id], ...], "evidences": [[subject entity, relation, object entity],...], "answer":answer}]

\section{Format Error}
\label{format_error}
\noindent\begin{minipage}{\textwidth}
  \begin{figure}[H]
    \includegraphics[width=\textwidth]{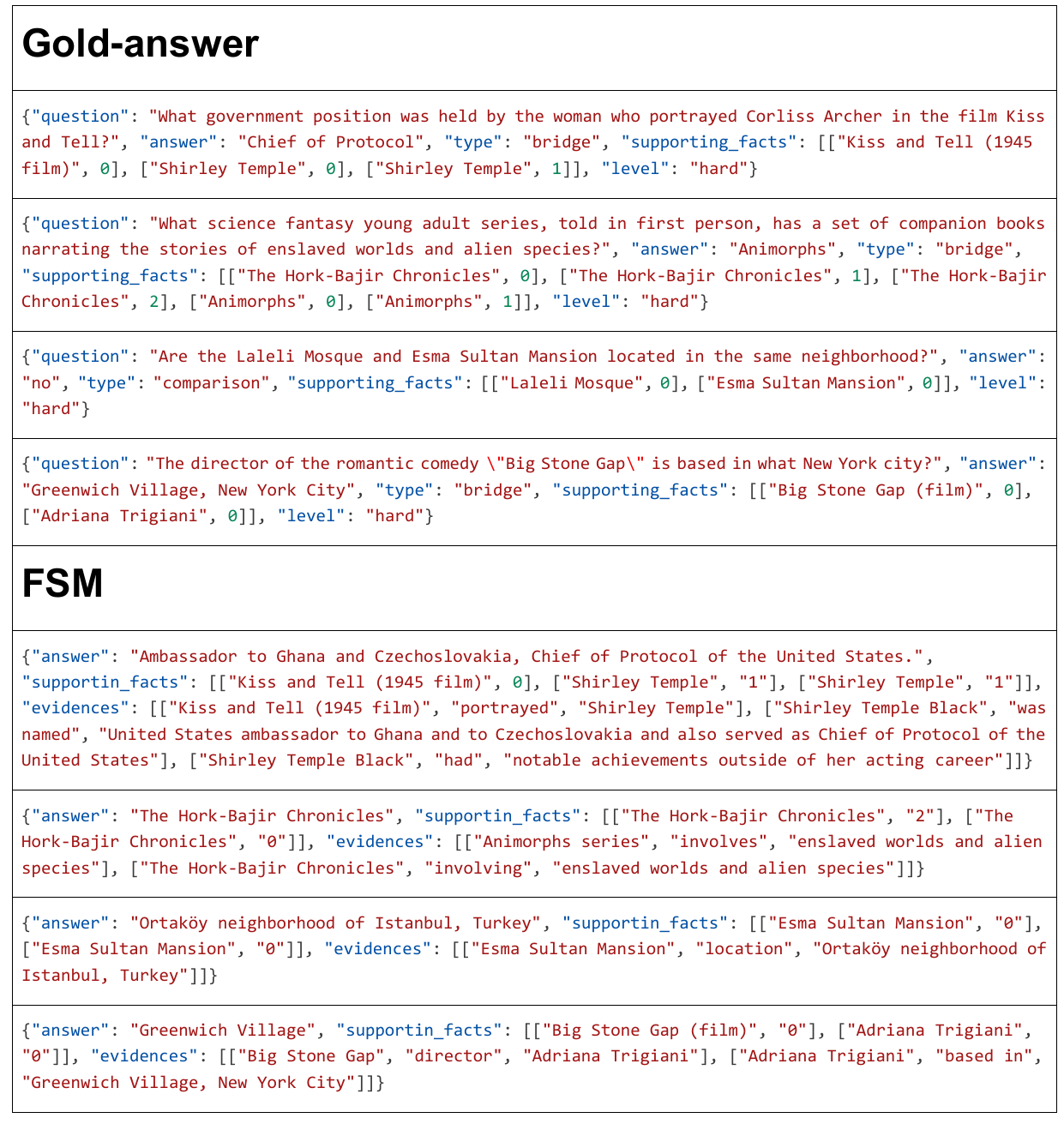}
    \caption{The outputs of SG-FSM are standard json format.}
    \label{format_error1}
  \end{figure}
\end{minipage}

\clearpage

\noindent\begin{minipage}{\textwidth}
  \begin{figure}[H]
    \includegraphics[width=\textwidth]{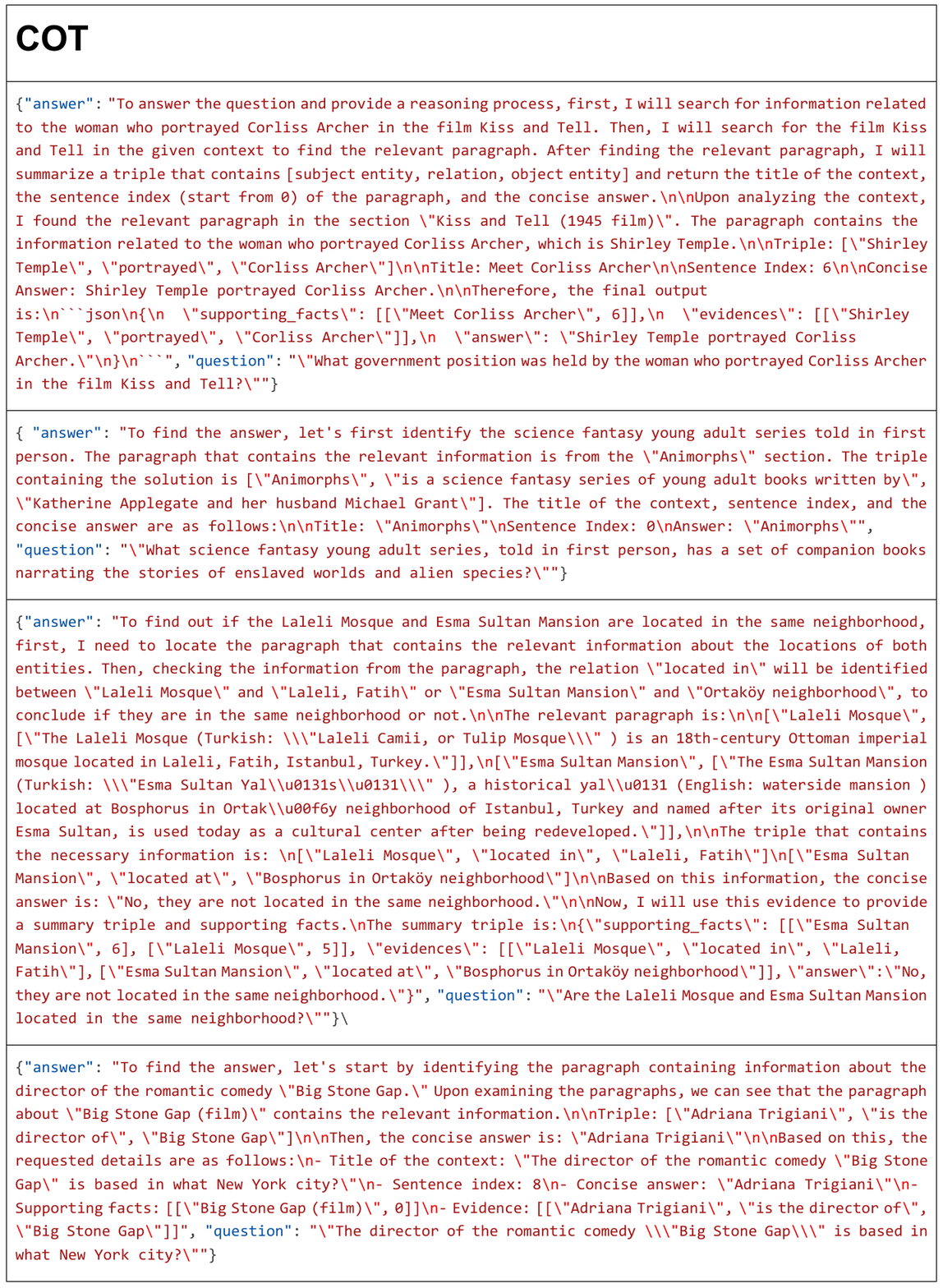}
    \caption{There are many error format examples for CoT. It is hard to get the answer for evaluation. }
    \label{format_error2}
  \end{figure}
\end{minipage}

\end{document}

%% file: floats/setting1.tex
\begin{table}[h]
\centering
 \small
\setlength{\tabcolsep}{4pt}
\resizebox{0.5\textwidth}{!}{
\renewcommand{\arraystretch}{1.2} 

\begin{tabular}{l l rrrrrrrrr} 
\toprule[1.2pt]
 & & \multicolumn{2}{c}{\textbf{ Musique}} & \multicolumn{2}{c}{ \textbf{HotpotQA}} & \multicolumn{2}{c}{\textbf{2WikiQA}} \\	\cmidrule(lr){3-4}\cmidrule(lr){5-6}\cmidrule(lr){7-8} 
 
 &  & EM  & F1   & EM   & F1   & EM   & F1  \\ \midrule
 
\multirow{5}{*}{\rotatebox[origin=c]{90}{\large GPT}} & \textbf{Direct} &  19.2 & 33.3 & 31.9 & 43.7 & 36.0 & 46.6    \\
&\textbf{CoT} & 20.6 & 35.6 & 32.1 & 45.5 & 38.1 & \textbf{53.0 }  \\
&\textbf{SP-CoT} & 14.4 & 28.4  & 24.8 & 37.4 & 23.2 & 36.0   \\

&\textbf{SG-FSM1} & 23.1 & 40.3 & 24.5 & 39.3 & 27.1 & 40.6 \\
&\textbf{SG-FSM2}  & \textbf{26.7} & \textbf{40.5} & \textbf{33.3} & \textbf{45.7} & \textbf{39.2} & 50.1 \\
 \midrule
\multirow{5}{*}{\rotatebox[origin=c]{90}{\large Qwen}} & \textbf{Direct} & 12.9 & 19.9  & 31.0 & 41.6 & 31.9 & 39.1   \\
&\textbf{CoT} & 14.1 & 24.0  & 30.6 & \textbf{42.7} & 39.9 & 49.8   \\
&\textbf{SP-CoT}  & 6.0 & 14.7 & 14.6 & 28.6 & 18.5 & 31.8 \\

&\textbf{SG-FSM1}  & \textbf{33.2}  & \textbf{48.5} & 28.0 &  37.4 & 39.1  & 47.9   \\
&\textbf{SG-FSM2}  & \textbf{33.2}  & \textbf{48.5} &  \textbf{32.2} & 41.3  &  \textbf{40.2} & \textbf{50.3 }  \\

 \bottomrule[1pt]
\end{tabular}
}
\caption{ Results on the MHQA benchmark by the GPT-3.5-turbo-1106 and Qwen-72B in setting 1 do not provide supporting evidence in the reasoning. }
\label{tab:setting1}
\end{table}

%% file: floats/bigdata.tex
\begin{table*}[ht]
\centering

\resizebox{\textwidth}{!}{
\renewcommand{\arraystretch}{1.2} 
\begin{tabular}{*{19}{c}}
\toprule[1.4pt]
 & \multirow{3}{*}{}  & \multicolumn{3}{c}{\textbf{\large Musique}} & \multicolumn{7}{c}{\textbf{\large HotpotQA}} & \multicolumn{7}{c}{\textbf{\large 2WikiQA}} \\	
 \cmidrule(lr){3-5}\cmidrule(lr){6-12}\cmidrule(lr){13-19} 
 & & \multicolumn{3}{c}{\normalsize  ans}  & \multicolumn{2}{c}{\normalsize ans} & \multicolumn{2}{c}{\normalsize sup} & \multicolumn{3}{c}{\normalsize joint} 
 & \multicolumn{2}{c}{\normalsize ans} & \multicolumn{2}{c}{\normalsize sup} & \multicolumn{3}{c}{\normalsize joint}   \\ 
 
 \cmidrule(lr){3-5}\cmidrule(lr){6-7} \cmidrule(lr){8-9} \cmidrule(lr){10-12} \cmidrule(lr){13-14} \cmidrule(lr){15-16} \cmidrule(lr){17-19}

~ &  & EM & F1  & Format & EM & F1 & EM & F1 & EM & F1 & Format &  EM & F1 & EM & F1 & EM & F1 & Format \\
\midrule
 \multirow{4}{*}{\rotatebox[origin=c]{90}{\large \textbf{Qwen}}}  & \textbf{Direct} & 18.2 & 30.9  & 84.0 & 31.6 & 42.8 & \textbf{2.6} & 26.4 & \textbf{1.3} & 13.4 & 90.7 & 6.7 & 8.0 & 1.6 & 5.5 & 1.0 & \textbf{2.6} & 89.8  \\ 
& \textbf{CoT} & 1.0 & 6.6  & 7.0 & 3.1 & 9.7 & 0.1 & 0.7 & 0.1 & 0.4 & 4.4 & 0.6 & 1.9 & 0 & 0.1 & 0.0 & 0.0 & 4.2   \\ 
& \textbf{SP-CoT} & 5.6 & 13.97  & 60.3 & 13.1 & 26.86 & 0.6 & 3.94 & 0.5 & 1.89 & 32.2 & 16.6 & 29.85 & 1.5 & 4.24 & 0.5 & 1.32 & 35.5   \\ 
& \textbf{SG-FSM1} & \textbf{26.2} & \textbf{41.2}  & \textbf{100.0}  & 22.5 & 33.3 & 0.7 & 9.9 & 0.4 & 3.6 & \textbf{100.0} & 27.6 & 37.9 & 4.7 & 25.8 & 1.9 & 9.1 & \textbf{100.0} \\ 
& \textbf{SG-FSM2} & 21.9 & 37.7  & \textbf{100.0} & \textbf{33.1} & \textbf{46.0} & 1.8 & \textbf{28.8} & 1.0 & \textbf{15.7} & \textbf{100.0} & \textbf{36.1} & \textbf{49.3} & \textbf{7.7}& \textbf{38.4} & \textbf{5.1} & \textbf{19.4} & \textbf{100.0}  \\ 

\midrule

\multirow{4}{*}{\rotatebox[origin=c]{90}{\large \textbf{GPT}}} & \textbf{Direct}  & 16.7 & 27.8 & 94.0 & \textbf{34.0} & \textbf{45.9} & 0.7 & 15.0 & 3.0 & 8.0 & 94.3 & \textbf{37.3} & \textbf{46.6} & 1.0 & 14.1 & \textbf{9.0} & 7.2 & 95.8 \\
& \textbf{CoT} & 4.5 & 13.6 & 14.7 & 12.3 & 26.0 & 0.4 & 4.5 & 2.0 & 17.8 & 16.2 & 8.2 & 19.3 & 0.2 & 1.3 & 1.0 & 4.6 & 7.0  \\

& \textbf{SP-CoT} & 14.33 & 27.92 & 84.9 & 24.0 & 36.48 & 2.56 & 24.52 & 1.1 & 10.11 & 82.8 & 23.11 & 36.03 & 9.56 & 34.87 & 3.0 & 12.50 & 79.2  \\

& \textbf{SG-FSM1}  & \textbf{26.0} & \textbf{38.4} & \textbf{100.0} & 23.4 & 32.0 & \textbf{2.4} & \textbf{29.3} & 2.0 & 9.8 & \textbf{100.0} & 30.1 & 40.0 & \textbf{14.2} & \textbf{47.0} & 2.0 & 8.5 & \textbf{100.0} \\
& \textbf{SG-FSM2} & 18.6 & 27.4  & \textbf{100.0} & 28.4 & 36.7 & 2.2 & 21.4 & \textbf{4.0} & \textbf{26.7} & \textbf{100.0} & 30.6 & 37.2 & 6.9 & 29.6 & 7.0 & \textbf{19.8} & \textbf{100.0}  \\
\bottomrule[1.4pt]
\end{tabular}
}
\caption{ Results on the MHQA benchmark by the GPT-3.5-turbo-1106 and Qwen-72B with zero-shot in setting 2, which requires providing the supporting evidence in the reasoning. "Ans" means answer. "Sup" means supporting paragraph index and title. "Joint" means evidence triples including relationshipS with sub-answers. }
\label{tab:setting2}
\end{table*}